\documentclass{article}
\usepackage{spconf,amsmath,graphicx}
\usepackage{amssymb}

\usepackage{bm}

\usepackage[sort&compress,numbers]{natbib}
\usepackage{siunitx} 

\usepackage{xspace}
\makeatletter
\DeclareRobustCommand\onedot{\futurelet\@let@token\@onedot}
\def\@onedot{\ifx\@let@token.\else.\null\fi\xspace}
 
\def\ie{i.e\onedot} 
 
 \def\vs{vs\onedot}

\makeatother

\newcommand{\Tstrut}[1][2.6ex]{\rule{0pt}{#1}}

\usepackage{enumitem}
\setlist{nosep, leftmargin=14pt}

\usepackage{textcomp}
\newcommand{\textapp}{\raisebox{0.5ex}{\texttildelow}}

\title{
From cells to survival: Hierarchical analysis of cell inter-relations in multiplex microscopy for lung cancer prognosis
}

\name{\begin{tabular}{c}%
 Olle Edgren Sch\"{u}llerqvist$^{\kern1pt\text{*1}}$,  
Jens Baumann$^{\kern0.5pt\text{*2}}$,
Joakim Lindblad$^{\kern1pt\text{1}}$,
Love Nordling$^{\kern1pt\text{3}}$,\\ 
Artur Mezheyeuski$^{\kern1pt\text{4,5}}$, 
Patrick Micke$^{\kern1pt\text{3}}$, 
Nata\v{s}a Sladoje$^{\kern1pt\text{1}}$%
\end{tabular}}
\address{%
$^{1}$ Department of Information Technology, SciLifeLab, Uppsala University, Uppsala, Sweden \\
$^{2}$ PAICON GmbH, Am Taubenfeld 21/2, 69123 Heidelberg, Germany \\
$^{3}$ Department of Immunology, Genetics and Pathology, Uppsala University, Uppsala, Sweden \\
$^{4}$ Molecular Oncology Group, Vall d'Hebron Institute of Oncology, Barcelona, Spain \\
$^{5}$ Vall d'Hebron Institute of Research, Barcelona, Spain
}%

\begin{document}

\maketitle
\def\thefootnote{*}\footnotetext{These authors contributed equally to this work.}\def\thefootnote{\arabic{footnote}}

\begin{abstract}
The tumor microenvironment (TME) has emerged as a promising source of prognostic biomarkers. To fully leverage its potential, analysis methods must capture complex interactions between different cell types. 
We propose HiGINE -- a hierarchical graph-based approach to predict patient survival (short \vs long) from TME characterization in multiplex immunofluorescence (mIF) images and enhance risk stratification in lung cancer. Our model encodes both local and global inter-relations in cell neighborhoods, incorporating information about cell types and morphology. Multimodal fusion, aggregating cancer stage with mIF-derived features, further boosts performance. 
We validate HiGINE on two public datasets, demonstrating improved risk stratification, robustness, and generalizability.
\end{abstract}

\section{Introduction}
\label{sec:intro}

Lung cancer has, worldwide, the highest incidence (12.4\%) and highest mortality (18.7\%) of all cancers~\cite{globocan2024}. Despite therapeutic advances, patient outcomes remain poor; improved understanding of disease mechanisms and refined risk stratification are essential for guiding treatment decisions. Current prognostic models for non-small cell lung cancer (NSCLC) rely mainly on clinical parameters such as tumor stage, performance status, and age, which fail to capture the biological complexity and diversity of NSCLC~\cite{hofman2025}. 

Over the last decade, the tumor microenvironment (TME) --  a network of immune and stromal cells interacting with tumor cells -- has emerged as a promising source of prognostic biomarkers.  Multiplex immunofluorescence (mIF) imaging enables simultaneous visualization of multiple proteins within tissue sections, providing high-resolution spatial insights into TME  composition and organization.  Characterization of immune cell densities and their spatial relationships in NSCLS has revealed consistent associations with recurrence-free and overall survival~\cite{backman2023,cai2025}.  However, recent studies indicate that moving beyond cell counts is critical to unlocking the full prognostic potential of the TME~\cite{sorin2023}.  

To capture the complex cellular architecture of TMEs, holistic and interpretable data-driven approaches are needed 
\cite{sorin2023}. Graph-based methods offer a framework to model cell neighborhoods and interactions~\cite{pati2022, zhou2019, Sextro2024}. In this study, we propose HiGINE -- a Hierarchical Graph Isomorphism Network with Edge features -- to encode both local and global relationships, while preserving biological interpretability by leveraging cell type information and morphological features. 

\smallskip\noindent\textbf{Our contributions are:} 
\begin{enumerate}[topsep=1pt plus 1pt minus 1pt]
    \item We combine mIF imaging with two-level graph-based modeling to capture local and global interactions within the TME in NSCLC and predict short \vs long survival.
    \item We demonstrate that edge-weighted graphs improve cell neighborhood encoding and model performance.
    \item We assess the benefit of multimodal fusion by incorporating cancer stage alongside mIF-derived features. 
    \item We validate our approach on two independent datasets (mIF and imaging mass cytometry (IMC)), confirming generalizability and competitive performance.
\end{enumerate}

\smallskip\noindent\textbf{Data and code availability:}
We publish complete code and dataset details at  https://github.com/MIDA-group/HiGINE\,.
\vspace*{-2pt}

\section{Background and previous work}
\label{sec:background}
\vspace*{-2pt}

mIF microscopy and IMC are established techniques for single-cell analysis within a spatial tissue context. Both enable simultaneous visualizations of multiple protein expression patterns: mIF typically detects \textapp10 fluorescence markers across entire tumor sections, while IMC captures \textapp40 proteins in smaller regions~\cite{eling2025,sorin2023}. Numerous studies, including ours~\cite{backman2023, mezheyeuski2023}, show that immunophenotyping based on mIF reveals prognostic associations between immune cell densities, spatial distributions, and clinical outcomes. 

However, a focus on individual cells -- localized and segmented in mIF images -- limits the ability to capture complex cell-cell interactions and their impact on patient survival. Recent approaches, instead, model cellular inter-relations using graph-based methods, representing cell neighborhoods and interaction networks within the TME~\cite{Sextro2024,pati2022,zhou2019,nakhli2023}. Graph neural networks (GNNs) enable hierarchical modeling of local (cell-level) and global (tissue-level) features~\cite{jimenezsanchez2022}. Recent studies propose using raw image data to extract discriminative features for patient stratification by learning-based approaches~\cite{atarsaikhan2025}. Multimodal strategies combining images of hematoxylin and eosin (H\&E) stained tissue with mIF, or IMC, further enhance predictive power~\cite{nakhli2023,schallenberg2025}. 

Although utilization of the complete raw image information offers strong predictive performance, it substantially increases the risk for so-called Clever Hans effects~\cite{vasquez2025detecting} and makes model interpretation considerably more complex. We emphasize the strong clinical value of interpretable models based on understandable features, such as cell locations, types, densities, and selected morphological properties. HiGINE offers competitive predictive performance together with transparency, crucial for trustworthy clinical adoption.

\section{Data}
\label{sec:Data}

We use two public datasets to develop and evaluate our proposed method. Although acquired by different techniques -- mIF and IMC -- both provide similar types of information about the TME in  NSCLC samples: spatial locations, types, and selected morphological features of the detected and segmented cells in the TME, along with extensive clinical metadata. Clinical parameters include follow-up time, a censoring indicator, the determined cancer stage, and routine patient data such as age, sex, and smoking status.

\medskip\noindent\textbf{Dataset 1:} A subset of the dataset used in \cite{backman2023}, consisting of 542 mIF images of tissue samples obtained from 298 patients with NSCLC. Details about sample preparation, imaging, cell segmentation, and feature extraction are provided in~\cite{backman2023}. 

\smallskip\noindent\textbf{Cell features:} 
The mean pixel intensity per each of 7 channels corresponding to the 4 immune cell markers FoxP3, CD4, CD20, CD8A, the cancer cell marker PanCK, the nuclear marker DAPI, and autofluorescence.  Morphological features: nucleus area (pixels), nucleus compactness (ratio of circumference to area), nucleus axis ratio (ratio of major/minor axes), and tissue category (tumor or stroma).

\smallskip\noindent\textbf{Binary survival label:} 
A threshold on the follow-up time is set to 1730 days after resection, splitting the cohort evenly into two groups of equal size (n=149). 

\smallskip\noindent\textbf{Binary cancer stage indicator:} 
The cancer stage (by 7th edition TNM system) splits patients into two groups, stage I (n=189) and stages II-IV (n=109).

\medskip\noindent\textbf{Dataset 2:} A public dataset \cite{sorin2023} consisting of single tissue spots obtained from 416 patients with adenocarcinoma of the lung. Each \qty{1.0}{mm^2} spot is stained with a 35-plex IMC panel, imaged at resolution of \qty{1}{\micro\meter}, and digitally processed to segment individual cells and extract 17 distinct cell phenotypes. 

\smallskip\noindent\textbf{Cell features:} 
For each cell, the one-hot encoded cell phenotype was extracted and used as the sole feature. 

\smallskip\noindent\textbf{Binary survival label:} 
A threshold on follow-up time of 36 months, as in \cite{sorin2023}, splits the cohort into two imbalanced groups (short-term: n=84, long-term: n=317), after removing censored patients from the short-term survival group. 

\smallskip\noindent\textbf{Binary cancer stage indicator:} 
The cancer stage (by 8th edition TNM system) splits patients into two groups (as suggested in \cite{sorin2023}), stages I-II (n=353) and stages III-IV (n=48).

\begin{figure}[t]
    \centering
    \includegraphics[width=0.9\linewidth]{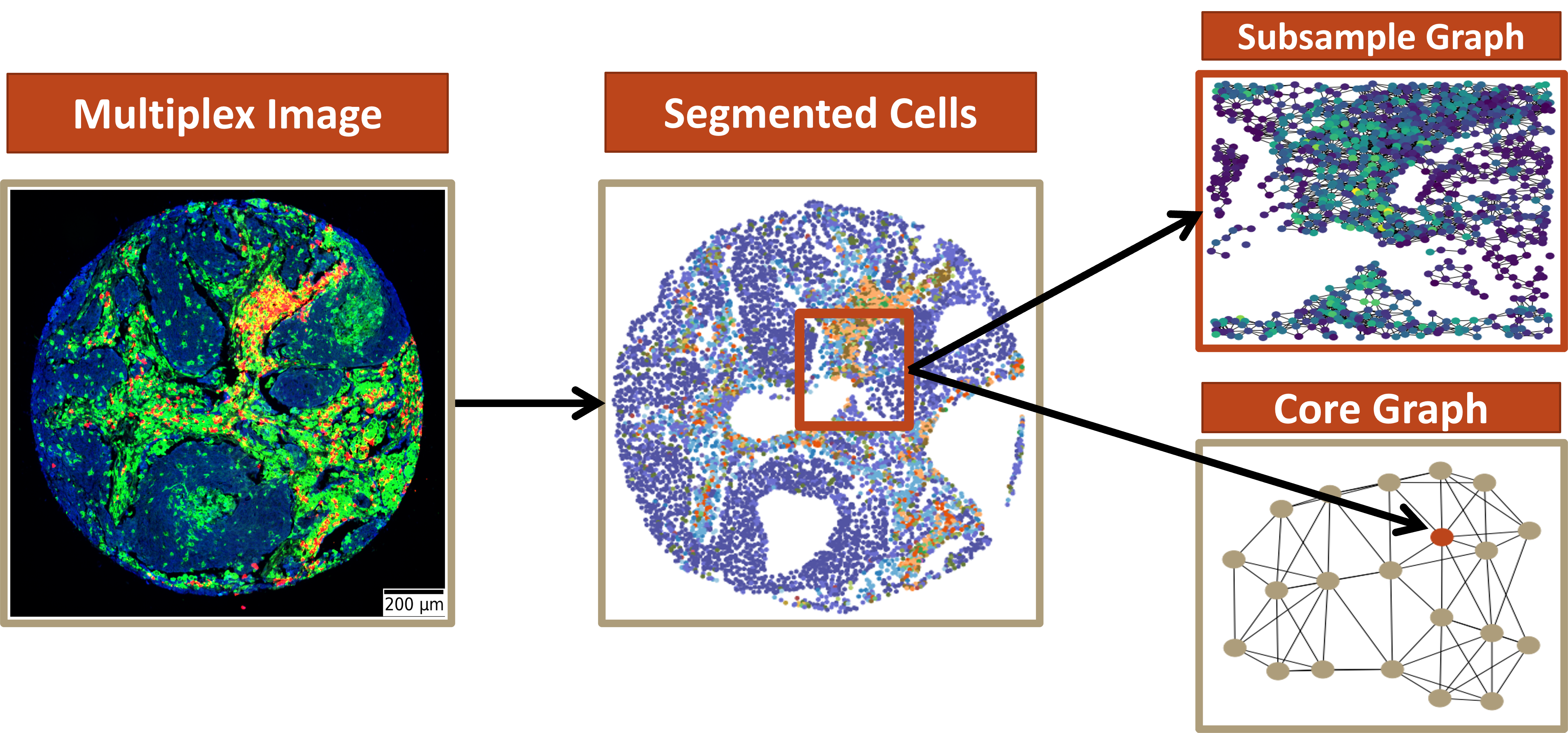}
    \caption{Each cell in the mIF image (7 channels, here visualized as RGB) is segmented, and its   quantitative features are extracted. The core is divided into subsample graphs using a sliding window approach. Each subsample graph is represented as a single node in the core graph.}
    \label{fig:graphs}
\end{figure}

\section{Method}
\label{sec:method}

HiGINE introduces a hierarchical two-level deep learning framework built on GNNs operating at different levels of granularity. At the  first level, a GNN encodes information from local cell neighborhoods, relying on local regions of cells sampled from the core. The second GNN aggregates these regional predictions to derive core-level predictions. 

\subsection{Graph Construction}
\label{ssec:graph}

We generate overlapping subsampled cell regions from each core using a sliding-window approach, maintaining a fixed number of cells (n=1000) in each region. Each subsample is represented as a graph, with the nucleus centroids as nodes, and with edges connecting neighboring cells within \qty{20}{\micro\meter} Euclidean distance (similar to~\cite{Barua2018}). Node attributes are the per-cell dataset-specific features described in Section~\ref{sec:Data}. As edge weights, we use the reciprocal of the Euclidean distance.

In the second stage, these subsample graphs are aggregated into core graphs: the geometric centroid of each subsample graph 
becomes a core graph node, whereas edges connect neighboring nodes at a maximum distance of \qty{330}{\micro\meter} 
(weighted as for the subgraphs). This process is illustrated in Figure~\ref{fig:graphs}. Core graph nodes are assigned learned features extracted by the network operating at the subsample-level.

\begin{figure}[t]
    \centering
    \includegraphics[width=0.98\linewidth]{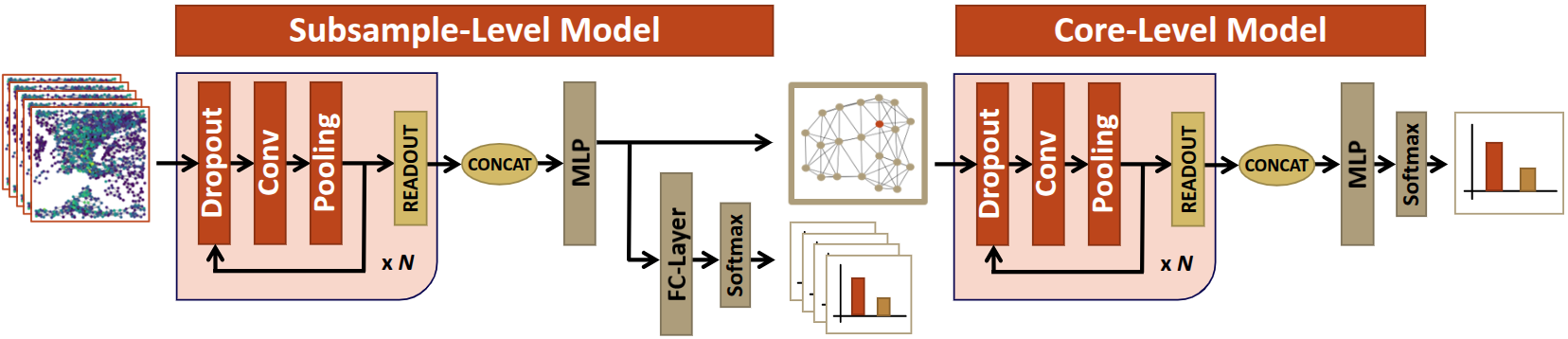}
    \caption{At the first level, the model classifies local cell graphs as short or long survival. Embeddings from the penultimate layer are used as node features of the graph at the second level, for core-level short- \vs long-term survival predictions.}
    \label{fig:modelArchitecture}
\end{figure}

\subsection{Model Architecture}
\label{ssec:Arch}

The general network design is shared between the models of the subsample and core levels. It consists of a graph convolutional network (GCN) as a graph encoder, followed by a multi-layer perceptron (MLP) as a prediction head. The GCN comprises multiple message-passing layers, each including a graph convolution and a self-attention graph (SAG) pooling operation~\cite{Lee2019}. We evaluate graph isomorphism network (GIN) convolutions~\cite{xu_how_2019-1}, both without and with edge features (GINE)~\cite{hu_strategies_2020}. For regularization, each graph convolution and linear layer is preceded by dropout operations.

\subsection{Training}
\label{ssec:Training}

HiGINE employs a two-step hierarchical training process. First, the subsample-level model is trained to predict binary survival labels from the local cell graphs. Features from the penultimate layer of this trained model are then propagated to the next level and assigned to nodes in the core-level graph. The core-level model uses  these enriched graphs to predict the survival at the core level.  Figure~\ref{fig:modelArchitecture} presents an overview. 

To improve robustness, we perform data augmentation by generating four sets of subsamples per core, with varying overlaps. The same augmentation strategy is used at test time, producing four predictions per core. All core-level predictions are finally aggregated to yield a single patient-level score, including predictions over multiple cores per patient in Dataset 1. Among several tested aggregation strategies, a simple average of softmax probabilities proved the most effective and interpretable.

Further technical details and hyperparameters are provided at https://github.com/MIDA-group/HiGINE\,.

\subsection{Multimodal Information Fusion }
\label{ssec:Fusion}

Cancer stage is one of the most important prognostic factors used in clinical practice. It is of interest to evaluate if it adds complementary information to the mIF data, and if the integration of the two can improve overall survival prediction.  

HiGINE supports multimodal fusion, enabling the inclusion of cancer stage information as a node feature in a graph. We evaluate such integration at the core-level.

\section{Evaluation}
\label{sec:eval}

Model performance is measured under a k-fold cross validation (CV) scheme to ensure robustness to data partitioning. HiGINE was evaluated over 10 folds in Dataset 1 and 5 folds in Dataset 2, each fold stratified w.r.t. clinical metadata. 

\subsection{Metrics} 
\label{ssec:metrics}

In each test fold of the CV loop, the model performance is measured using the area under the receiver operating characteristic curve (AUROC) and the concordance index (c-index). The AUROC is calculated using the binary survival label and the predicted positive class softmax probability. The c-index is calculated using the predicted risk score (corresponding to the negative class probability), the observed survival time, and the censoring indicator.

\subsection{Baselines}
\label{ssec:Baseline}

We benchmark HiGINE against selected baseline approaches. 

\smallskip\noindent{\bf Upper bound:} HiGINE is trained to predict the binary survival label (see Section~\ref{sec:Data}). Using the ground-truth label as the risk score, we derive an {\it upper bound} for the c-index (under our training conditions).

\smallskip\noindent{\bf Cancer stage:} We report c-index when using (only) the binary cancer stage as the risk score.  

\smallskip\noindent{\bf Shallow learning:} We evaluate performance of the support vector classifier (SVC) and logistic regression (LogReg), trained on the mean and standard deviation of the per-cell features within each core. For Dataset 1, we further split the  distribution of cells and their associated features into tumor and stroma regions. Thus, SVC and LogReg rely solely on summary statistics and cannot leverage spatially resolved cellular interactions.  

\smallskip\noindent{\bf GNN-based approach}: Following~\cite{Sextro2024} and \cite{Schnake2022}, we evaluate the performance of a recent end-to-end GIN approach (utilizing the implementation by \cite{Yuan2023}). Here, a single GNN is used for survival prediction from cell-graphs (over the entire cores). The GNN consists of a 3-layer GIN without graph pooling followed by a 2-layer MLP. Multiple cores per patient are processed separately, and output logits are averaged. Input graphs have the same nodes as used for HiGINE, and edges are defined according to a 3-nearest neighbor (3-NN) algorithm. The cancer stage is optionally fused to the graph embedding before the MLP. A 5-fold nested CV scheme is used for hyperparameter tuning during training.

\begin{table}
  \centering
  \caption{Performance of HiGINE on the two considered datasets, in terms of AUROC and c-index (higher is better, avg$\pm$std).  1st row: Upper bound of a binary predictor. 2nd row: Prediction from the binary cancer stage (CS) only.}\vspace{0.4em}
  \label{tab:main-results}
  \small
  \resizebox{\columnwidth}{!}{
  \begin{tabular}{@{}l@{\:}|@{\;}c@{\:}|@{\;}c@{\;\;}c@{\,}|@{\;}c@{\;\;}c@{\,}}
   \multicolumn{2}{c}{} & \multicolumn{2}{c}{\textbf{Dataset 1}} & \multicolumn{2}{c}{\textbf{Dataset 2}} \\[0.1ex]
   \textbf{Method} & \textbf{CS} & \textbf{AUROC} & \textbf{C-Index}  & \textbf{AUROC} & \textbf{C-Index}\\\hline
   \Tstrut Label & $-$ & $1$ & $0.783$ & $1$ & $0.732$ \\
   $-$ & $\checkmark$ & $0.624$ & $0.588$ & $0.605$ & $0.568$ \\\hline
   \Tstrut SVC & $-$ & $0.577\scriptscriptstyle\pm0.12$ & $0.552\scriptscriptstyle\pm0.09$ & $0.603\scriptscriptstyle\pm0.06$ & $0.556\scriptscriptstyle\pm0.06$ \\
   LogReg & $-$ & $0.566\scriptscriptstyle\pm0.13$ & $0.546\scriptscriptstyle\pm0.09$ & $0.609\scriptscriptstyle\pm0.03$ & $0.568\scriptscriptstyle\pm0.03$ \\
   GIN & $-$ & $0.549\scriptscriptstyle\pm0.08$ & $0.555\scriptscriptstyle\pm0.05$ & $0.573\scriptscriptstyle\pm0.06$ & $0.541\scriptscriptstyle\pm0.05$  \\
   HiGINE & $-$ & $\bm{0.649}\scriptscriptstyle\pm0.08$ & $\bm{0.591}\scriptscriptstyle\pm0.04$ & $\bm{0.660}\scriptscriptstyle\pm0.07$ & $\bm{0.588}\scriptscriptstyle\pm0.03$ \\
   \hline
   \Tstrut SVC & $\checkmark$ & $0.616\scriptscriptstyle\pm0.11$ & $0.588\scriptscriptstyle\pm0.09$ & $0.674\scriptscriptstyle\pm0.04$ & $0.610\scriptscriptstyle\pm0.04$ \\
   LogReg & $\checkmark$ & $0.632\scriptscriptstyle\pm0.12$ & $0.597\scriptscriptstyle\pm0.09$ & $0.663\scriptscriptstyle\pm0.04$ & $0.605\scriptscriptstyle\pm0.03$ \\
   GIN & $\checkmark$ & $0.637\scriptscriptstyle\pm0.04$ & $0.614\scriptscriptstyle\pm0.04$ & $0.679\scriptscriptstyle\pm0.08$ & $0.617\scriptscriptstyle\pm0.05$ \\
   HiGINE & $\checkmark$ & $\bm{0.690}\scriptscriptstyle\pm0.11$ & $\bm{0.617}\scriptscriptstyle\pm0.07$ & $\bm{0.703}\scriptscriptstyle\pm0.06$ & $\bm{0.632}\scriptscriptstyle\pm0.04$
   \end{tabular}} 
\end{table}

\section{Results}
\label{sec:results}

HiGINE outperforms all evaluated baselines on both datasets. Table~\ref{tab:main-results} summarizes the results. The first row indicates the upper bound on performance, derived from the ground-truth labels. The second row shows predictions based on the binary cancer stage (CS) only. Results indicate that HiGINE extracts highly relevant information from the cell graphs constructed on TMEs, outperforming CS as a predictor. Of the alternatives, only LogReg on Dataset 2 succeeds in that. Furthermore, HiGINE clearly extracts information from the cell graphs which is  complementary to CS, since HiGINE with integrated CS outperforms, by a wide margin, both the model without CS fusion and the CS-only predictor. 

\begin{figure}[tb]
    \centering
    \includegraphics[width=0.98\linewidth]{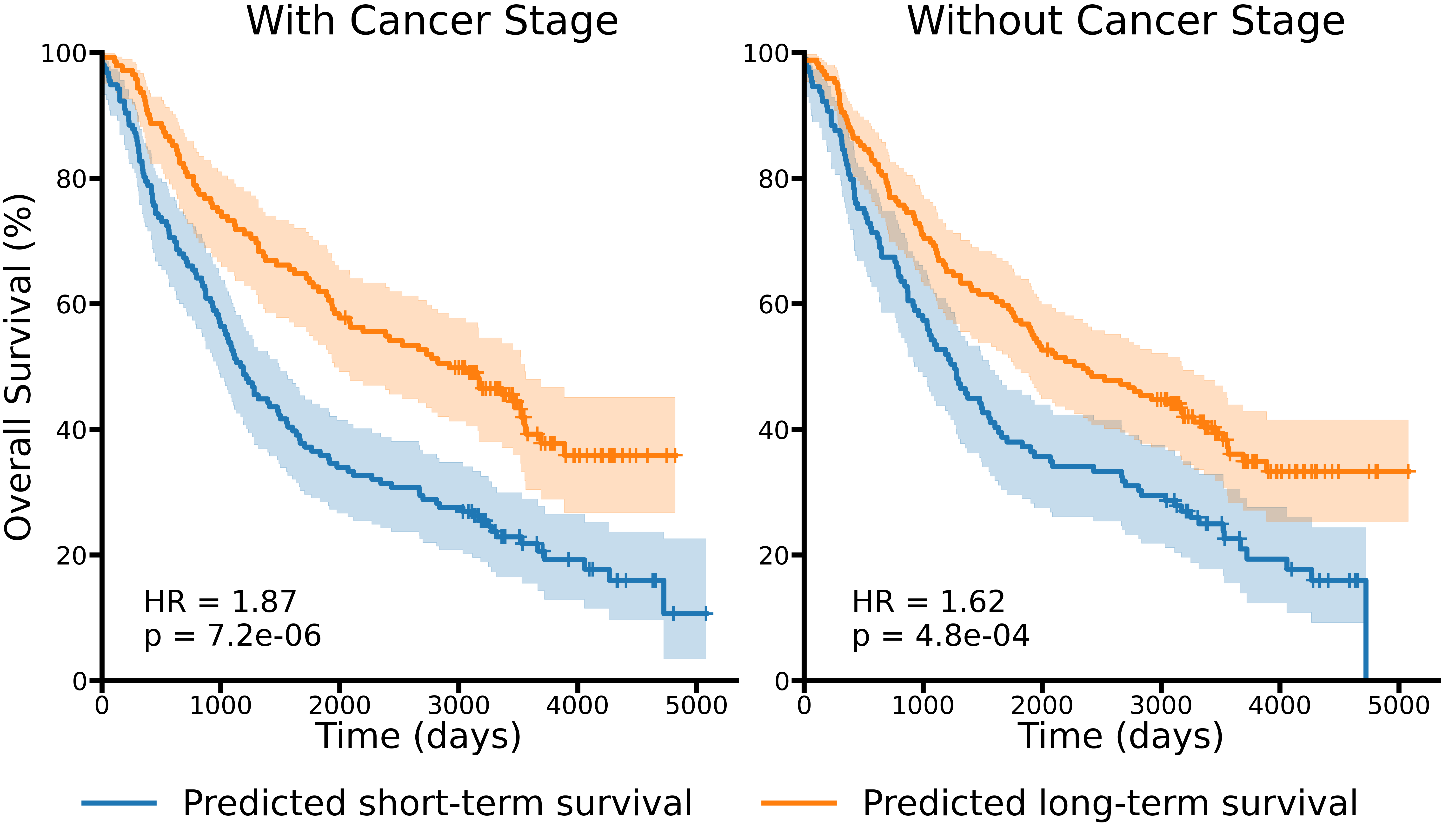}
    \caption{Kaplan-Meier curves for patients from Dataset~1 split into short/long survival, predicted by HiGINE when (left) incorporating CS in the model input, and (right) relying on mIF image data alone. The $p$-values are based on the log-rank test.}
    \label{fig:km_plot}
\end{figure}

Kaplan-Meier analysis of patient stratification on Dataset~1 by HiGINE (Fig.~\ref{fig:km_plot}) demonstrates that the model effectively separates patients into risk groups with significantly different survival outcomes. Log-rank test $p$-values for separation of curves are $7.2\cdot10^{-6}$  and $4.8\cdot10^{-4}$, when based on HiGINE predictions with, and without, CS. Corresponding hazard ratios (HR) computed from a Cox proportional hazards model are $1.87$ and $1.62$.

\subsection{Ablation Study}

 To evaluate the relevance of the individual components of HiGINE, we perform a comprehensive ablation study. The results are summarized in Table~\ref{tab:ablation-results}. 

\smallskip\noindent\textbf{Edge weighting}: We consider weighted and non-weighted edges (\ie, GIN \vs GINE) in the constructed cell graphs. The column denoted as \textbf{'E'} (Table~\ref{tab:ablation-results}) indicates whether the model uses edge weights or not. We observe that weighted edges generally have a positive impact on performance. 

\smallskip\noindent\textbf{Graph hierarchy}: The column denoted as \textbf{'Hi'} (Table~\ref{tab:ablation-results}) indicates if the core-level network is used, or alternatively, if the patient-level prediction is derived directly from the subsampled graphs. We observe a positive impact of the hierarchical approach on both datasets.  

\smallskip\noindent\textbf{Cancer stage information:}
The column denoted as \textbf{'CS'} (Table~\ref{tab:ablation-results}) indicates if the binary CS is used by the core-level network. We observe that the fusion of cell-graph and CS information yields the best performance.

\begin{table}[tb]
  \centering
  \caption{Ablation study. Performance of HiGINE on the two considered datasets, in terms of AUROC and c-index (higher is better, avg$\pm$std). \textbf{'E'}: w/wo edge-weighting. \textbf{'Hi'}: w/wo 2nd hierarchical stage. \textbf{'CS'}: w/wo cancer stage fusion.}\vspace{0.4em}
  \label{tab:ablation-results}
  \small
  \resizebox{\columnwidth}{!}{
  \begin{tabular}{@{\;}c@{\;}|@{\:}c@{\:}|@{\,}c@{\,}|@{\;\,}c@{\;\;}c@{\,}|@{\;\,}c@{\;\;}c@{\,}}
   \multicolumn{3}{c}{} & \multicolumn{2}{c}{\textbf{Dataset 1}} & \multicolumn{2}{c}{\textbf{Dataset 2}} \\[0.1ex]
   \textbf{E} & \textbf{Hi} & \textbf{CS} & \textbf{AUROC} & \textbf{C-Index} & \textbf{AUROC} & \textbf{C-Index} \\\hline
   \Tstrut $-$ & $-$ & $-$ & $0.606\scriptscriptstyle\pm0.08$ & $0.558\scriptscriptstyle\pm0.06$ & $0.610\scriptscriptstyle\pm0.06$ & $0.562\scriptscriptstyle\pm0.03$ \\
   $\checkmark$ & $-$ & $-$ & $0.634\scriptscriptstyle\pm0.09$ & $0.577\scriptscriptstyle\pm0.05$ & $0.605\scriptscriptstyle\pm0.05$ & $0.568\scriptscriptstyle\pm0.03$ \\
   $-$ & $\checkmark$ & $-$ & $0.627\scriptscriptstyle\pm0.06$ & $0.577\scriptscriptstyle\pm0.04$ & $0.621\scriptscriptstyle\pm0.07$ & $0.568\scriptscriptstyle\pm0.03$  \\
   $\checkmark$ & $\checkmark$ & $-$ & $\bm{0.649}\scriptscriptstyle\pm0.08$ & $\bm{0.591}\scriptscriptstyle\pm0.04$ & $\bm{0.660}\scriptscriptstyle\pm0.07$ & $\bm{0.588}\scriptscriptstyle\pm0.03$ \\
   \hline
   \Tstrut $-$ & $\checkmark$ & $\checkmark$ & $0.680\scriptscriptstyle\pm0.10$ & $0.610\scriptscriptstyle\pm0.06$ & $0.654\scriptscriptstyle\pm0.09$ & $0.599\scriptscriptstyle\pm0.03$ \\
   $\checkmark$ & $\checkmark$ & $\checkmark$ & $\bm{0.690}\scriptscriptstyle\pm0.11$ & $\bm{0.617}\scriptscriptstyle\pm0.07$ & $\bm{0.703}\scriptscriptstyle\pm0.06$ & $\bm{0.632}\scriptscriptstyle\pm0.04$
  \end{tabular}}
\end{table}

\section{Discussion and Conclusion}
\label{sec:conclusion}

HiGINE -- a hierarchical graph isomorphism network with edge features -- effectively models cell neighborhoods in the TME and integrates multimodal data to improve risk stratification in NSCLC. Unlike shallow learning and non-hierarchical graph approaches, HiGINE successfully extracts prognostic information from cell interactions, both with and without the integration of clinical stage data. While hierarchical subsampling, together with our data augmentation strategies (multiple subsamplings), enhance data utilization and stabilize learning, 
the observed high variance in results indicates that larger training datasets would be beneficial.

\pagebreak
\clearpage

\section*{Compliance with Ethical Standards}
\label{sec:ethics}

This research study was conducted retrospectively using human subject data made available in open access by \cite{backman2023,sorin2023}. Ethical approval was not required as confirmed by the license attached with the open access data.

\section*{Conflicts of Interests}
\label{sec:acknowledgments}
We report no conflicts of interests. J. Baumann was with Uppsala University when working on this project. 
This work is supported by Swedish Cancer Society projects 22 2353 Pj and 22 2357 Pj, the Swedish Research Council grant 2022-03580, and the SciLifeLab \& Wallenberg DDLS Program (grant KAW2024.0159). Computations were facilitated by the Berzelius resource provided by the Knut and Alice Wallenberg Foundation at the National Supercomputer Centre.

\bibliographystyle{IEEEbib}
\bibliography{refs}

\end{document}